\pdfoutput=1

\documentclass[11pt]{article}

\usepackage{naacl2021}

\usepackage{times}
\usepackage{latexsym}

\usepackage[T1]{fontenc}

\usepackage[utf8]{inputenc}

\usepackage{microtype}

\usepackage{amsmath,amssymb,amsfonts,bm}
\usepackage{subfigure}
\usepackage[linesnumbered,ruled, vlined]{algorithm2e}
\usepackage{multirow}
\usepackage{color}
\usepackage{booktabs}
\usepackage{graphicx}

\title{Rethinking Network Pruning--- \\under the Pre-train and Fine-tune Paradigm}

\author{Dongkuan Xu\textsuperscript{\rm 1} \and Ian En-Hsu Yen\textsuperscript{\rm 2} \and Jinxi Zhao\textsuperscript{\rm 2} \and Zhibin Xiao\textsuperscript{\rm 2} \\ 
\textsuperscript{\rm 1}The Pennsylvania State University, State College, PA, USA \\
\textsuperscript{\rm 2}Moffett AI, Los Altos, CA, USA \\
\{dux19\}@psu.edu, \{ian.yan, jx.zhao, zb.xiao\}@moffett.ai
}

\begin{document} 
\maketitle
\begin{abstract}
Transformer-based pre-trained language models have significantly improved the performance of various natural language processing (NLP) tasks in the recent years. While effective and prevalent, these models are usually prohibitively large 
for resource-limited deployment scenarios. 
A thread of research has thus been working on applying network pruning techniques under the pretrain-then-finetune paradigm widely adopted in NLP. However, the existing pruning results on benchmark transformers, such as BERT, are not as remarkable as the pruning results in the literature of convolutional neural networks (CNNs). In particular, common wisdom in pruning CNN states that sparse pruning technique compresses a model more than that obtained by reducing number of channels and layers \cite{elsen2020fast,zhu2017prune}, while existing works on sparse pruning of BERT yields inferior results than its small-dense counterparts such as TinyBERT \cite{jiao2020tinybert}. In this work, we aim to fill this gap by studying how knowledge are transferred and lost during the pre-train, fine-tune, and pruning process, and proposing a knowledge-aware sparse pruning process that achieves significantly superior results than existing literature.
We show for the first time that sparse pruning compresses a BERT model significantly more than reducing its number of channels and layers. Experiments on multiple data sets of GLUE benchmark show that our method outperforms the leading competitors with a 20-times weight/FLOPs compression and neglectable loss in prediction accuracy\footnote{The code and models are publicly available at \url{https://github.com/DerronXu/SparseBERT}. The work was done during an internship at Moffett AI.}.


\end{abstract}

\section{Introduction}
\label{sec:intro}
Pre-trained language models, such as BERT~\cite{devlin2019bert}, become the standard and effective methods for improving the performance of a variety of natural language processing (NLP) tasks. 
These models are pre-trained in a self-supervised fashion and then fine-tuned for supervised downstream tasks. 
However, these models suffer from the heavy model size, making them impractical for resource-limited deployment scenarios and incurring cost concerns~\cite{strubell2019energy}. 

In parallel, an emerging subfield has studied the redundancy in deep neural network models~\cite{zhu2017prune,gale2019state} and proposed to 
prune networks without sacrificing performance, such as the lottery ticket hypothesis~\cite{frankle2019}. 
Common wisdom in CNN literature shows that sparse pruning leads to 
more compression rate than structural pruning. For example, for the same number of parameters (0.46M), the sparse MobileNets improve by 11.2\% accuracy over the dense ones~\cite{zhu2017prune}.
However, similar conclusions are not observed for pre-trained language models.

The main question this paper attempts to answer is: how to perform sparse pruning under the pre-train and fine-tune paradigm? 
Answering this question correctly is challenging. 
First, these models adopt pre-training and fine-tuning procedures, during which the general-purpose language knowledge and the task-specific knowledge are learned respectively. Thus, it is desirable and challenging to keep the weights that are important to both knowledge during pruning. Second, unlike CNNs, pre-trained language models have a complex architecture consisting of embedding, self-attention, and feed-forward layers.

\begin{figure*}[!t]
\vspace{-0cm}
\centering
  \subfigure[General Pre-Training \& Fine-Tuning]{\label{fig:prune_comp1}
  \includegraphics[width=0.4\textwidth,height=0.14\textwidth]{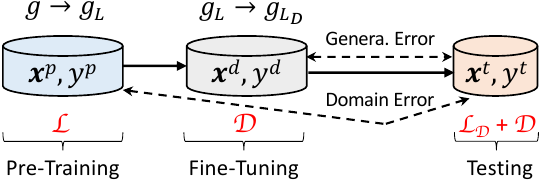}
  }\hspace{2cm}
  \subfigure[Pruning at Fine-Tuning]{\label{fig:prune_comp2}
  \includegraphics[width=0.4\textwidth,height=0.14\textwidth]{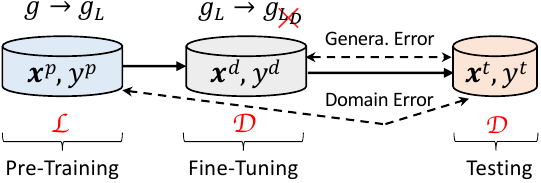}
  }\hspace{0cm}
  \subfigure[Pruning at Pre-training]{\label{fig:prune_comp3}
  \includegraphics[width=0.4\textwidth,height=0.14\textwidth]{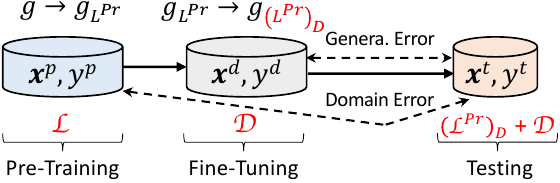}
  }\hspace{1cm}
  \subfigure[Pruning at Distilling (Proposed)]{\label{fig:prune_comp4}
  \includegraphics[width=0.5\textwidth,height=0.16\textwidth]{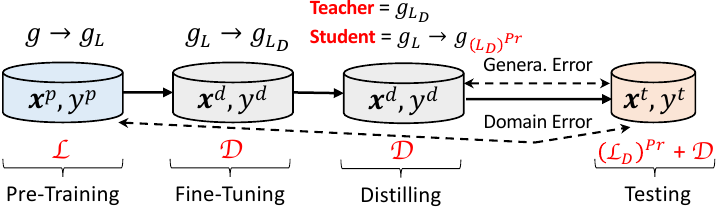}
   }\vspace{0cm}
\caption{How knowledge is transferred under different pruning strategies. (a) is the general pre-training and fine-tuning procedure (Section~\ref{sec:knowledgetransfer}). 
$g$ is an encoder. $g_L$ and $g_{L_D}$ are the encoders well-trained on the pre-training and fine-tuning datasets respectively. $\mathcal{L}$ and $\mathcal{D}$ are the general-purpose language knowledge and the task-specific knowledge respectively. There is a domain error between pre-training and testing, and a generalization error between fine-tuning and testing. (b) and (c) are two basic pruning strategies (Section~\ref{sec:knowledgeaware-basic-prune}).
Both $\mathcal{L}_D$ and $\mathcal{L}^{pr}$ are subsets of knowledge $\mathcal{L}$. $\mathcal{L}_D$ is related to the downstream task. $\mathcal{L}^{pr}$ is preserved in a pruned encoder $g_{L^{pr}}$. (d) is the proposed pruning strategy (Sections~\ref{sec:knowledgeaware-proposed-prune}-\ref{sec:knowledgeaware-proposed-distill}). $(\mathcal{L}^{pr})_D$ refers to the knowledge obtained by first pruning and then fine-tuning. $(\mathcal{L}_D)^{pr}$ corresponds to first fine-tuning and then pruning while distilling.}
\label{fig:prune_comp_1}\vspace{-0.2cm}
\end{figure*}

\begin{figure}[!t]
\vspace{-0cm}
\centering
  \subfigure[Knowledge Relationship]{\label{fig:prune_comp5}
  \includegraphics[width=0.25\textwidth,height=0.1\textwidth]{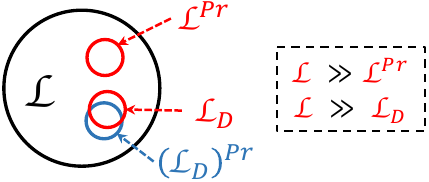}
   }\vspace{0cm}
  \subfigure[Why Distilling]{\label{fig:prune_comp6}
  \includegraphics[width=0.2\textwidth,height=0.09\textwidth]{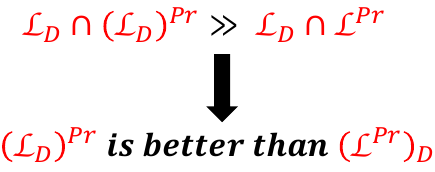}
   }\vspace{0cm}
\caption{Knowledge Analysis.}
\label{fig:prune_comp_2}\vspace{-0.2cm}
\end{figure}

To address these challenges, we propose SparseBERT, a knowledge-aware sparse pruning method for pre-trained language models, with a special focus on the widely used BERT model. SparseBERT is executed in the fine-tuning stage. It preserves both general-purpose and task-specific language knowledge while pruning.
To preserve the general-purpose knowledge learned during pre-training, SparseBERT uses the pre-trained BERT without fine-tuning as the initialized model and prunes the linear transformations 
in self-attention and feed-forward layers, which is inspired by the recent findings that self-attention and feed-forward layers are overparameterized~\cite{michel2019sixteen,voita2019analyzing} and are also the most computation consumption parts~\cite{ganesh2020compressing}. 
To learn the task-specific task knowledge 
during pruning while preserving the general-purpose knowledge at the same time, we apply knowledge distillation~\cite{hinton2015distilling}. We adopt the task-specific fine-tuned BERT as the teacher network and the pre-trained BERT that is being pruned as the student. We feed the downstream task data into the teacher-student framework to train the student to reproduce the behaviors of the teacher.

We summarize different types of BERT pruning approaches in Figure~\ref{fig:prune_comp_1} (see Section~\ref{sec:knowledgeaware} for detailed discussion) 
Experimental results on the GLUE benchmark 
demonstrate that SparseBERT outperforms all the leading competitors and achieves 1.4\% averaged loss with down to only 5\% remaining weights 
compared to BERT-base.
\section{Related Work}
\label{sec:related}
A lot of efforts have been made on studying network redundancy and pruning networks without accuracy loss~\cite{gale2019state,Renda2020Comparing}. For example, the work on lottery ticket hypothesis~\cite{frankle2019} showed that there exist sparse smaller subnetworks capable of training to full accuracy in CNNs. 
Common wisdom in CNN literature shows that spare pruning leads to much more compression rate than structural pruning~\cite{gale2019state,elsen2020fast}. For example, for the same number of parameters (0.46M), the sparse MobileNets achieve 61.8\% accuracy while the dense ones achieve 50.6\%~\cite{zhu2017prune}.
However, similar observations are not observed in existing approaches for pre-trained language models~\cite{fan2019reducing,michel2019sixteen,chen2020lottery,mccarley2020structured,jiao2020tinybert}.
Our method aims to fill the gap and summarize these pruning strategies. 
There are other compression approaches for pre-trained language models, such as quantization~\cite{zafrir2019q8bert} and weight factorization~\cite{wang2019structured}, which are out of the scope of this work. 

\section{SparseBERT}
\label{sec:sparsebert}
We first formalize the knowledge transfer involved in fine-tuning pre-trained language models. 
Then, we introduce our SparseBERT. 

\begin{figure*}[!t]
\begin{center}
\centerline{\includegraphics[width=1\textwidth,height=0.475\textwidth]{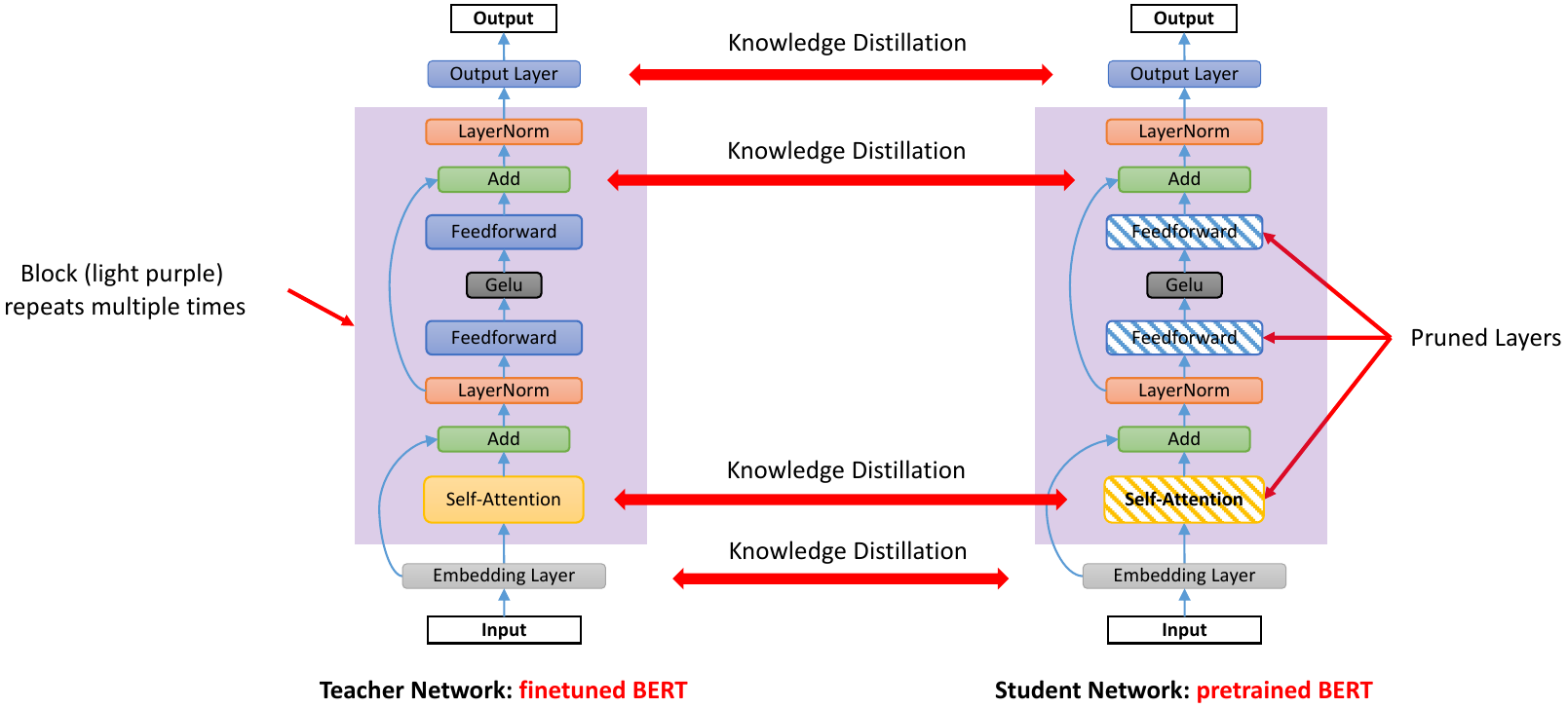}}
\caption{Illustration of the proposed knowledge-aware compression. Pruning is performed in parallel with distillation, based on specific data from downstream tasks.}
\label{fig:prune-distill}
\end{center}
\end{figure*}

\subsection{Knowledge Transfer under the Pre-train and Fine-tune Paradigm}
\label{sec:knowledgetransfer}
The practice of fine-tuning pre-trained language models has become prevalent in various NLP tasks.
The two-stage procedure is illustrated in Figure~\ref{fig:prune_comp1}.
The language model is denoted by $f$ = $g$ $\circ$ $h$, where $g$ is a text encoder and $h$ is a task predictor head.
Text encoders, like Transformers in BERT, are used to map input sentences to hidden representations and task predictors further map the representations to the label space. The pre-trained model is trained on a large amount of data examples $(\mathbf{x}^p,y^p)$ from the pre-training task domain via different tasks that resemble language modeling. 

During pre-training, the general-purpose language knowledge, denoted by $\mathcal{L}$, is learned based on $(\mathbf{x}^p,y^p)$. $\mathcal{L}$ contains a subset that is related to the downstream task, denoted by $\mathcal{L}_D$, and the amount of $\mathcal{L}$ is far greater than that of $\mathcal{L}_D$ (see Figure~\ref{fig:prune_comp5}). 
To transfer knowledge $\mathcal{L}$ (especially $\mathcal{L}_D$) from pre-training domain to downstream domain, the well-trained encoder $g_L$ is used to initialize the downstream encoder. 
In fine-tuning, downstream encoder is trained based on the task-specific knowledge $\mathcal{D}$ preserved in a small amount of data examples $(\mathbf{x}^d,y^d)$ from downstream domain.
Finally, the well-trained downstream encoder $g_{L_D}$ is evaluated on test data.

\subsection{Knowledge-Aware Compression}
\label{sec:knowledgeaware}

\subsubsection{Two Basic Pruning Strategies}
\label{sec:knowledgeaware-basic-prune}
Intuitively, there are two pruning strategies. One is that pruning is applied to the downstream encoder $g_L$ during fine-tuning 
(see Figure~\ref{fig:prune_comp2}). However, because the loss to update the weights during fine-tuning is exclusively based on the data examples $(\mathbf{x}^d,y^d)$ from the downstream task domain, this pruning strategy might destruct the knowledge $\mathcal{L}_D$, which is learned based on $(\mathbf{x}^p,y^p)$ and encoded in the initialization of $g_L$. 

The other strategy is that pruning is executed during pre-training (see Figure~\ref{fig:prune_comp3}). The generated pruned network preserves a subset of knowledge $\mathcal{L}$, denoted by $\mathcal{L}^{pr}$. Unfortunately, because this strategy ignores the downstream task information and the amount of $\mathcal{L}$ is extremely large, i.e., $\mathcal{L}$ $\gg$ $\mathcal{L}^{pr}$, 
the knowledge $\mathcal{L}^{pr}$ could be much different from $\mathcal{L}_D$ that we hope to preserve (see Figure~\ref{fig:prune_comp5}).

\subsubsection{The Proposed Pruning Strategy}
\label{sec:knowledgeaware-proposed-prune}
As shown in Figure~\ref{fig:prune_comp4}, 
SparseBERT executes pruning at the distilling stage.
It prunes the pre-trained encoder without fine-tuning, $g_L$, while fine-tuning the pruned encoder based on the downstream dataset $(\mathbf{x}^d,y^d)$. 
Recent findings indicate that self-attention and feed-forward layers are overparameterized and are the most computation consumption parts~\cite{michel2019sixteen,voita2019analyzing,ganesh2020compressing}. Thus, SparseBERT applies network pruning to the linear transformations matrices in self-attention and feed-forward layers (see Figure~\ref{fig:prune-distill}). 
The choice of pruning approach is flexible. We choose magnitude weight pruning~\cite{han2015learning} in this paper, mainly because it is one of the most effective and popular pruning methods. More details about the pruning strategy used in SparseBERT can be found in the codes.

\subsubsection{Knowledge Distillation Helps Pruning Preserve Task-Specific Knowledge}
\label{sec:knowledgeaware-proposed-distill}
To mitigate the loss of $\mathcal{L}_D$, 
we propose to utilize knowledge distillation while pruning. We use the task-specific fine-tuned BERT as the teacher network and the pre-trained BERT that is being pruned as the student (see Figure~\ref{fig:prune_comp4} and Figure~\ref{fig:prune-distill}). 
The motivation is that the task-specific fine-tuned BERT preserves $\mathcal{L_D}$. By feeding downstream task data $(\mathbf{x}^d,y^d)$ into the teacher-student framework, we 
help the student reproduce the behaviors of the teacher to learn both $\mathcal{L}_d$ and $\mathcal{L}$ as much as possible.

We design the distillation loss as
\begin{equation}
    L_{distil} = L_{emb} + L_{att} + L_{hid} + L_{prd}.
\end{equation}
$L_{emb}$ = MSE($\mathbf{E}^S,\mathbf{E}^T$) is the difference between the embedding layers of student and teacher.
$L_{att}$ = $\sum$MSE($\mathbf{A}^S_i,\mathbf{A}^T_i$) is the difference between attention matrices and $i$ is the layer index.
$L_{hid}$ = $\sum$MSE($\mathbf{H}^S_i,\mathbf{H}^T_i$) is the difference between hidden representations.
$L_{prd}$ = -softmax($\mathbf{z}^T$) $\cdot$ log\_softmax($\mathbf{z}^S/$temp) is the soft cross-entropy loss between the logits of student and teacher. temp represents the temperature value. The proposed distillation loss is inspired by~\cite{jiao2020tinybert} and it helps the student imitate the teacher's behavior as much as possible. In addition, we perform the same data augmentation as \cite{jiao2020tinybert} does to generate more task-specific data for teacher-student learning. Notably, the choices of distillation loss and data augmentation method are flexible and we found the ones we adopted worked well in general.

\section{Experiments}
\label{sec:exp}


\subsection{GLUE Benchmark}
\label{sec:glue}
We evaluate 
SparseBERT on four data sets from the GLUE benchmark~\cite{wang2018glue}. To test if SparseBERT is applicable across tasks, we include the tasks of both single sentence and sentence-pair classification. We report the results on dev sets. We run 3, 20, 20, 50 epochs for QNLI, MRPC, RTE, CoLA separately. 
The baselines include BERT-base, ELMo~\cite{peters2018deep}, BERT-PKD~\cite{sun2019patient}, Bert-of-Theseus~\cite{xu2020bert}, DistilBERT~\cite{sanh2019distilbert}, MiniLM~\cite{wang2020minilm}, TinyBERT~\cite{jiao2020tinybert}, BERT-Tickets~\cite{chen2020lottery}, CompressBERT~\cite{gordon2020compressing}, and RPP~\cite{guo2019reweighted}.

The results are shown in Table~\ref{tb:glue_res}. Compared to BERT-base, SparseBERT achieves 1.4\% averaged performance loss with down to 5\% weights. In addition, SparseBERT outperforms all leading competitors with the highest sparsity.


\begin{table}[!ht]
\scriptsize
\addtolength{\tabcolsep}{-2pt}
\renewcommand{\arraystretch}{1}
\centering
\vspace{-0cm}
\begin{tabular}{lcccccc}
\toprule
\multirow{2}*{Method} & Remain. & QNLI & MRPC  & RTE &  CoLA & \multirow{2}*{Avg.}\\
                     & Weights & (Acc) & (F1) & (Acc)& (Mcc) & \\
\midrule
\multicolumn{7}{c}{\textit{Without Pruning}} \\
BERT-base       & -       & 91.8  & 88.6  & 69.3  &  56.3  & 76.5 \\
ELMo            & -       & 71.1  & 76.6  & 53.4  &  44.1  & 61.3 \\
\midrule
\multicolumn{7}{c}{\textit{Structural Pruning}} \\
BERT$_6$-PKD    & 50\%    & 89.0  & 85.0  & 65.5  &  45.5  & 71.3 \\
BERT-of-Theseus & 50\%    & 89.5  & 89.0  & 68.2  &  51.1  & 74.5 \\
DistilBERT      & 50\%    & 89.2  & 87.5  & 59.9  &  51.3  & 72.0 \\
MiniLM$_6$      & 50\%    & 91.0  & 88.4  & 71.5  &  49.2  & 75.0 \\
TinyBERT$_6$    & 50\%    & 90.4  & 87.3  & 66.0  &  54.0  & 74.4 \\
TinyBERT$_4$    & 18\%    & 88.7  & 86.8  & 66.5  &  49.7  & 72.9 \\
\midrule
\multicolumn{7}{c}{\textit{Sparse Pruning}} \\
BERT-Tickets    & 30-50\% & 88.9  & 84.9  & 66.0  &  53.8  & 73.2 \\
CompressBERT & 10\%    & 76.8  & -     & -     &  -     & -    \\
RPP             & 11.6\%  & 88.0  & 81.9  & 67.5  &  -     & -    \\
SparseBERT      & 5\%     & 90.6  & 88.5  & 69.1  &  52.1  & 75.1 \\
\bottomrule
\end{tabular}
\caption{Comparison on the dev sets of GLUE.}\vspace{0cm}
\label{tb:glue_res}
\vspace{-0.2cm}
\end{table}

\subsection{SparseBERT v.s. Pruning at Downstream}
We compare SparseBERT with the pruning described in Figure~\ref{fig:prune_comp2} on the question answer tasks of SQuAD v1.1 and v2.0~\cite{rajpurkar2016squad,rajpurkar2018know}. Given a question and a passage containing the answer, the two tasks are to predict the answer text span in the passage. The difference between them is that SQuAD v2.0 allows for the possibility that no short answer exists in the passage. We follow the general setting of SparseBERT, except that we only apply the logit distillation, i.e., $L_{distil}$ = $L_{prd}$, and do not perform data augmentation, which are the most common distillation strategies.

\begin{figure}[!th]
\vspace{-0in}
\centering
  \subfigure[SQuAD v1.1.]{\label{fig:visual-1}
  \includegraphics[width=0.22\textwidth,height=0.18\textwidth]{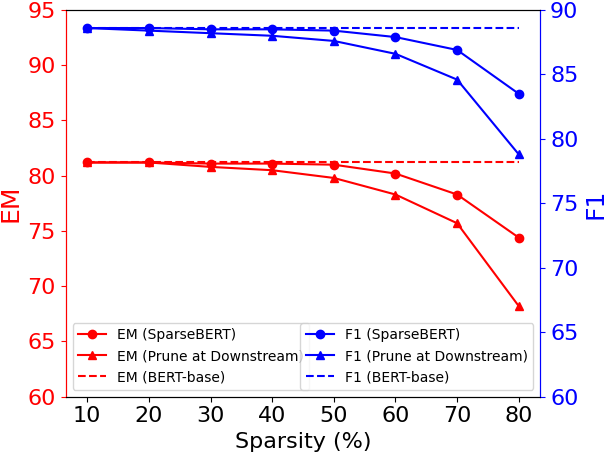}
  }\vspace{0cm}
  \subfigure[SQuAD v2.0.]{\label{fig:visual-2}
  \includegraphics[width=0.22\textwidth,height=0.18\textwidth]{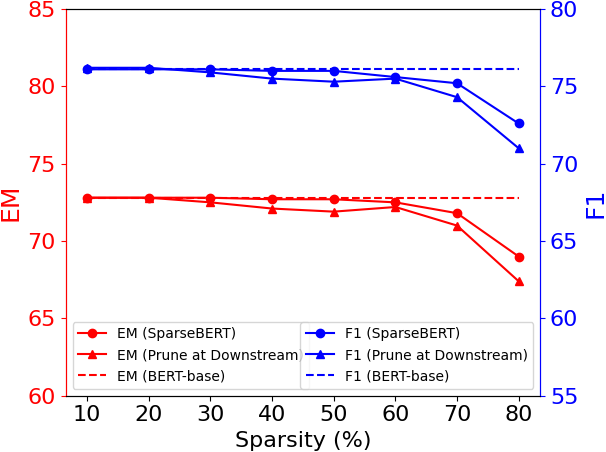}
  }\vspace{-0.2cm}
\caption{Performance comparison of SparseBERT and the pruning approach described in Figure~\ref{fig:prune_comp2}.}
\label{fig:squad}\vspace{-0.3cm}
\end{figure}

The results are shown in Figure~\ref{fig:squad}. It is observed that SparseBERT consistently outperforms the baseline method, especially at high sparsity. The performance gain of SparseBERT decreases on SQuAD v2.0 mainly because SQuAD v2.0 is more challenging than SQuAD v1.1. These observations demonstrate advantage of SparseBERT compared to pruning at downstream.

\subsection{SparseBERT v.s. Pruning at Pre-Training}
To get more insights about the advantage of SparseBERT over the pruning described in Figure~\ref{fig:prune_comp3}, we compare their fitting abilities. Specifically, we use TinyBERT as an example of the baseline pruning method. We compare SparseBERT with TinyBERT with 4 layers and 312 hidden dimensions, which has a similar number of parameters as SparseBERT (sparsity=95\%). SparseBERT only distills knowledge from the same layers as TinyBERT does.

\begin{figure}[!th]
\vspace{-0in}
\centering
  \subfigure[Loss (training set).]{
  \includegraphics[width=0.22\textwidth,height=0.17\textwidth]{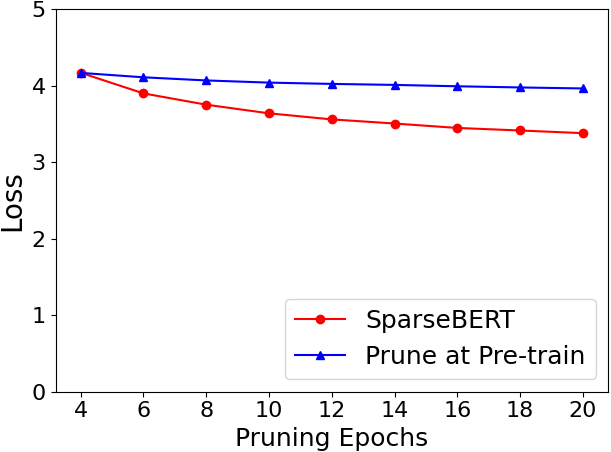}
  }\vspace{0cm}
  \subfigure[Performance (dev set).]{
  \includegraphics[width=0.22\textwidth,height=0.17\textwidth]{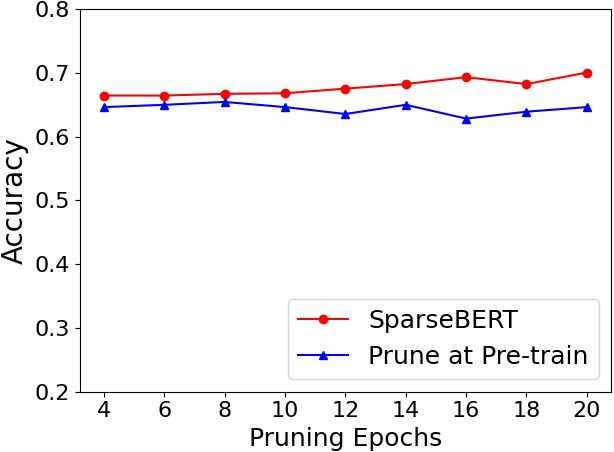}
  }\vspace{-0.2cm}
\caption{Fitting ability comparison of SparseBERT and the pruning approach described in Figure~\ref{fig:prune_comp3}.}
\label{fig:fitting}\vspace{-0.2cm}
\end{figure}

We vary the number of pruning epochs and report the results (loss on training set and accuracy on dev set) on RTE in Figure~\ref{fig:fitting}. It is observed that SparseBERT consistently shows smaller training loss while higher evaluation performance, which demonstrates that SparseBERT has a better fitting ability when pruning compared to the baseline.

\section{Discussion}
\label{sec:dis}

\subsection{Hardware Performance}

\begin{figure}[!t]
\begin{center}
\centerline{\includegraphics[width=0.34\textwidth,height=0.23\textwidth]{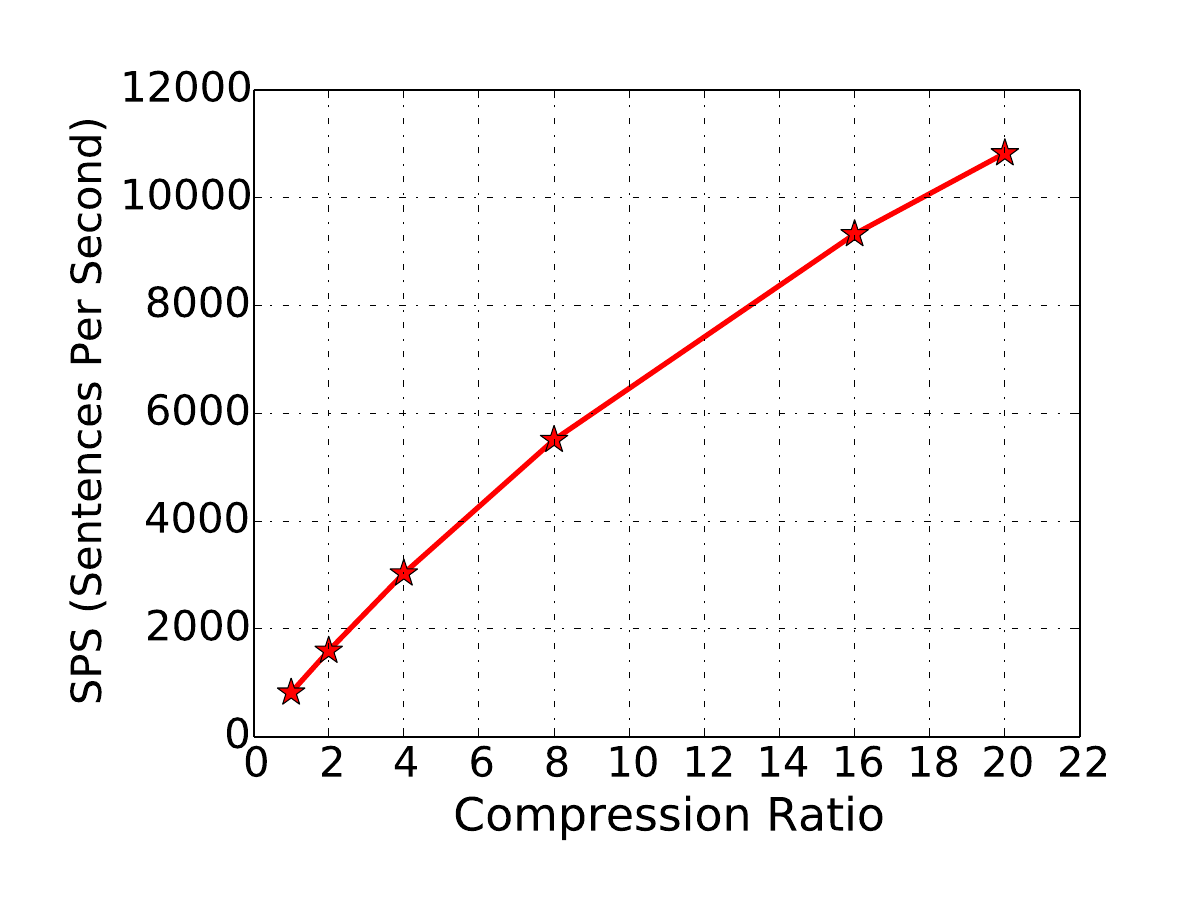}}
\caption{Hardware performance under different compression ratios on the MRPC dataset, with 818, 1594, 3029, 5508, 9326, and 10826 SPS (sentences per second) respectively.}
\label{fig:hw_perf}
\end{center}
\vspace{-0.4cm}
\end{figure}

Sparse networks were not hardware-friendly in the past. However, hardware platforms with sparse tensor operation support have been rising up. For example, the latest release of Nvidia high-end GPU A100 has native support of sparse tensor operation up to 2x compression rate, while startup company such as Moffett AI has developed computing platform with sparse tensor operation acceleration up to 32x compression rate.

Here we deployed SparseBERT of different sparse compression ratios (1, 2, 4, 8, 16, 20) on Moffett AI's latest hardware platform ANTOM to measure the real inference speedup induced by sparse compression, where `4' indicates the model is compressed by a factor of 4, with 75\% of the parameters being zeros. As shown in Figure~\ref{fig:hw_perf}, the sparse compression has almost linear speedup up to 4x and leads to more than 10x speedup when compression rate is 20x.

\subsection{Reduction of Parameters and FLOPS}
We studied the reduction of parameters and FLOPS. For example, on the MRPC dataset, BERT-base (backbone) vs SparseBERT (backbone) = 85.53 vs 4.84 (\#parameters, M) and BERT-base vs SparseBERT = 10.87 vs 0.54 (GFLOPS).

\subsection{Inference/Training Time}

We studied the time and convergence speed. For example, to get the reported 20x pruned result (Table~\ref{tb:glue_res}), it needed 12 epochs of fine-tuning on MRPC and each epoch took 1.5 h (two RTX 2080 Ti). The inference time was around 20 s.

\section{Conclusion}
\label{sec:conclusion}

We introduce SparseBERT, a knowledge-aware sparse pruning method for pre-trained language models, with a focus on BERT. We summarize different types of BERT pruning approaches and compare SparseBERT with leading competitors. Experimental results on GLUE and SQuAD benchmarks demonstrate the superiority of SparseBERT.
\section{Acknowledgements}
\label{sec:acknowledgements}
We thank Xiaoqi Jiao for his valuable discussion and feedback on this work.

\clearpage

\bibliography{anthology}
\bibliographystyle{acl_natbib}


\end{document}